\documentclass[lettersize,journal]{IEEEtran}
\usepackage{amsmath,amsfonts}
\usepackage{algorithm}
\usepackage{algpseudocode}
\usepackage{bm}
\usepackage{array}
\usepackage[caption=false,font=normalsize,labelfont=sf,textfont=sf]{subfig}
\usepackage{textcomp}
\usepackage{stfloats}
\usepackage{url}
\usepackage{verbatim}
\usepackage{graphicx}
\usepackage{cite}

\usepackage{listings}
\usepackage{fancyvrb}
\usepackage[inline]{enumitem}
\usepackage[thinlines]{easytable}
\usepackage{multirow}
\usepackage{array}
\usepackage[export]{adjustbox}

\usepackage{amsmath,amsfonts}
\usepackage{graphicx}
\usepackage{xcolor}
\usepackage{tikz}
\usepackage{amsfonts}
\usepackage{dsfont}

\usetikzlibrary{positioning}

\definecolor{cyan1}{RGB}{64,224,208}
\definecolor{cyan2}{RGB}{72,209,204}
\definecolor{blue1}{RGB}{65,105,225}
\definecolor{purple1}{RGB}{148,0,211}
\definecolor{yellow1}{RGB}{218,165,32}
\definecolor{orange1}{RGB}{255,140,0}
\definecolor{red1}{RGB}{220,20,60}

\newcolumntype{C}[1]{>{\centering\let\newline\\\arraybackslash\hspace{0pt}}m{#1}}

\begin{document}

\title{LEED: Local Embedding Evolution Distance for over-smoothing estimation and virtual node selection in GNN}

\author{
\centering
\IEEEauthorblockN{Killian Cressant, Pedro B. Velloso,~\textit{CNAM, France}}

\thanks{Killian Cressant and Pedro B. Velloso are from the CEDRIC Lab, Conservatoire National des Arts et M\'etiers (Cnam), 75003 Paris, France (email: killian.cressant@cnam.fr; pedro.velloso@cnam.fr).
}
}

\maketitle

\begin{abstract}

Graph Neural Networks (GNNs) suffer from two fundamental limitations: over-smoothing, where node representations become indistinguishable with depth, and over-squashing, where long-range information is compressed through limited message-passing channels. Existing metrics such as Dirichlet energy provide global characterizations of over-smoothing but lack the resolution to analyze node-level behavior and guide architectural improvements.
In this paper, we propose LEED (Local Embedding Evolution Distance), a novel local metric that quantifies over-smoothing by tracking the evolution of individual node embeddings across layers. By operating at the node level, LEED enables fine-grained analysis of representation dynamics during training, revealing heterogeneous over-smoothing patterns that are invisible to global energy-based measures. This locality induces informative node importance scores, interpreted as embedding-driven centrality measures.
We leverage LEED to design a more efficient strategy for virtual node selection. Unlike existing approaches that depend on multiple heuristic centrality measures, our method uses LEED as a unique criterion to guide the construction of Local Virtual Nodes to mitigate over-squashing.
Experiments show that LEED provides more informative diagnostics than Dirichlet energy while preserving global evaluation, and enables more effective virtual node integration, improving GNN performance across datasets.

\end{abstract}

\begin{IEEEkeywords}
GNN, graph rewiring, over-smoothing, over-squashing, virtual node
\end{IEEEkeywords}

\section{Introduction}.
\IEEEPARstart{G}{raph}  Neural Networks (GNN) have been developed in the past years to leverage structured topological data to enhance the performance of neural network models. They have been successfully applied across a wide range of domains, including  networking applications~\cite{network, new_related}, where network topology naturally defines the graph structure, molecular datasets such as classical ENZYMES and MUTAG, citation networks with CORA dataset~\cite{GCN} and even traditional  computer vision tasks~\cite{computer_v}. Despite its recent success, GNNs still face fundamental challenges specially regarding the influence of the message-passing process on the quality and expressiveness of node embeddings~\cite{express}. 

In this context, one important issue in GNNs is the over-smoothing problem, which makes node embeddings indistinguishable as information is propagated across the graph, leading to a severe performance degradation~\cite{survey_oversm}.
Another important challenge is the over-squashing problem~\cite{oversquashing.25}. It occurs specially  when messages from distant nodes are aggregated  as they propagate across graph layers, causing structurally important information to be compressed, distorted, or discarded. This problem is intensified in the presence of bottleneck nodes and limited-capacity embeddings.

Previous studies have shown that using too many layers in GNN architectures inevitably results in over-smoothing even if it is necessary to pass long-range dependencies.To mitigate this issue, Kipf and Welling proposed using a shallow two-layer GCN~\cite{GCN}. Subsequent works, such as~\cite{SGCN}, further demonstrated that the GCN model can be simplified by removing intermediate nonlinear activation functions without compromising its performance. This phenomenon relating the number of layers to the loss of expressive power was later studied in~\cite{expressive}.
In addition, different approaches have been developed to combat over-smoothing~\cite{journal1}. In particular, researchers have proposed to modify the global loss function by incorporating regularization terms~\cite{gradrew}. In this direction, a prominent approach involves the use of Dirichlet energy~\cite{dirichlet}. By monitoring or regularizing this energy, one can prevent the signal from collapsing into a uniform state.

More recently, the focus has shifted toward graph rewiring~\cite{doc}. This technique relies on the fact that the original graph structure, often dictated by physical or logical relationships in the data, is not necessarily the optimal topology for training GNNs~\cite{gstruct}. Therefore, graph rewiring techniques dynamically~\cite{dyna2} or pre-emptively alter the edge or node set to improve message flow, effectively bypassing bottlenecks to reduce over-squashing~\cite{squash} or pruning redundant paths to prevent over-smoothing.
A central challenge in these frameworks lies in identifying the critical nodes that would most benefit from the addition of local transformations. 
However, existing graph rewiring approaches largely rely on empirically testing a range of classical centrality measures, such as degree, betweenness or PageRank, and selecting the one that yields the best performance.

In this paper we propose LEED (Local Embedding Evolution Distance), a node-level metric designed to characterize over-smoothing by monitoring how individual node representations change across network layers. Therefore, the main advantage of LEED consists of addressing both the over-smoothing and over-squashing problems simultaneously. 
By focusing on per-node dynamics, LEED captures fine-grained variations in representation collapse that global energy-based indicators fail to detect. In addition, we exploit its local properties to replace all the classical centrality metrics to find the critical nodes for graph rewiring. Therefore, the centrality metric is no longer a parameter of the model.
Our experimental results demonstrate that LEED offers more precise and informative diagnostics of over-smoothing than Dirichlet energy while maintaining comparable global assessment capabilities. Furthermore, we apply LEED as a unique metric to guide two different graph rewiring solutions to select the critical nodes. Finally, we show that LEED improves GNN performance across six classical datasets~\cite{tudata} with different characteristics by mitigating over-squashing, and, at the same time, controlling over-smoothing.

The remainder of this paper is organized as follows. In Section~\ref{sec:back}, we present our preliminary notation related to GNNs, as well as the basic aspects of graph rewiring.
Section~\ref{sec:rw} presents the related work concerning the over-smoothing estimation and critical nodes selection in graph rewiring. In Section~\ref{sec:leed} we introduce LEED along with its design issues and its definitions. Section~\ref{sec:cns} we define practical usage of critical nodes. In Section~\ref{subsection:experiment} and~\ref{sec:results}, we present our experiments and the main results. Finally, Section~\ref{sec:conc} presents our conclusions and future work.

\section{Background}
\label{sec:back}

Let $G=(V,E)$ be a undirected graph with a set of node $V$ and a set of edges $E$. We consider the adjacency matrix of the graph $\mathbf{A}$, the neighborhood of a node $i$ is $N_i=\{j\in V | (j,i) \in E\}$ and the embedding input features is $\mathbf{X}=(X_1,...,X_n)$, with $n=|V|$. Then, the general equation of a layer of a GNN (working with message-passing layer) can be written as:

$$ X_i^{(k+1)}= Update^{(k)}(X_i^{(k)},Agg^{(k)}(\{X_j^{(k)}, j \in N_i\})) $$

If we consider the GCN of Kipf and Welling \cite{GCN}, we obtain:

 $$ \mathbf{X}^{(k+1)}= \sigma (\tilde{\mathbf{D}}^{-\frac{1}{2}}\tilde{\mathbf{A}}\tilde{\mathbf{D}}^{-\frac{1}{2}} \mathbf{X}^{(k)}\mathbf{W^{(k)}})  $$
with  $ \tilde{\mathbf{A}}= \mathbf{A} +\mathbf{I_N}$, $A$ the adjacency matrix of the graph and $\mathbf{I_N}$ the identity matrix, $\tilde{\mathbf{D}}$ the diagonal matrix elements of $\tilde{\mathbf{A}}$, $\mathbf{W^{(k)}}$ the trainable weight matrix of the GNN and $\sigma$ an activation function, such as ReLU.
This equation comes from the initial Graph Convolutional Networks (GCN) paper~\cite{GCN}, where they develop a GNN based on graph convolution. The main idea is to write the convolution in the Fourier domain, then use a Chebyshev polynomial approximation of the eigenbase of the Laplacian $\mathbf{\Delta}=\mathbf{I_N}-\mathbf{D}^{-\frac{1}{2}}\mathbf{A}\mathbf{D}^{-\frac{1}{2}}$ to  finally construct the layer operation, creating a first-order spectral filter layer, followed by an activation function. The activation function is the update function and the aggregation is made by convolution. Since GCNs inherit a strong influence from the GNN message-passing paradigm, this equation remains quite accurate for most GNN models, with the matrix formulation preserved and only minor modifications to the update rule~\cite{SGCN} or to the aggregation mechanism, as in~GAT~\cite{gat}.

\subsection{Over-smoothing vs over-squashing}
\label{sec:over-over}
In the last years, these two phenomena have been observed in GNNs and have attracted significant attention in the literature~\cite{survey_oversm,  oversquashing.25, mad}. The basic definitions for these two problems are as follows:

\begin{itemize}
    \item \textbf{Over-smoothing:} a result of repeated aggregations that cause node features to converge to a stationary point, effectively washing out the local signals necessary for downstream tasks.
    \item  \textbf{Over-squashing:} this occurs when a structural bottleneck (such as a bridge or ``bottleneck" node) is forced to compress a disproportionately large volume of information from one subgraph to another. This forced compression leads to a significant loss of information, preventing long-range dependencies from being effectively captured.
\end{itemize}

While over-smoothing typically arises from excessive message-passing across densely connected structures, leading to indistinguishable node representations, over-squashing occurs under the opposite topological conditions.
As a consequence, a trade-off emerges when attempting to mitigate both phenomena simultaneously~\cite{over_over}. Hence, in general, these two problems are addressed separately in the literature.

Over-smoothing is usually handled through regularization or GNN layer modifications. Graph rewiring, on the other hand, is predominantly used to alleviate over-squashing, and the most common techniques include:
\begin{itemize}
    \item Adding virtual nodes~\cite{VN}. It consists of adding one~\cite{vn1} or several central virtual nodes~\cite{multivn} to the original graph. As the over-squashing problem intensifies  when the commute time between nodes is high, the key idea is to reduce drastically and efficiently the graph diameter, and as a consequence, the commute time of the graph.
    \item Using topological curvature~\cite{log2}, such as Ricci curvature~\cite{curvature1}, Olivier curvature or similar measures. In this case, the main goal is to add edges in specific regions of the graph that act as topological bottlenecks, thereby alleviating structural constraints.
    \item Lastly, some works focus on expander graphs. Expander graph is a family of graphs that has a high Cheeger constant, which can be associated with fewer topological bottlenecks. In this setting, virtual nodes and topological modifications are combined to construct an entirely new graph~\cite{cayley}.
    These approaches share the same theoretical motivation as curvature-based methods: modifying or removing all bottleneck from the graph structure will improve the modeling of long-range dependencies.
\end{itemize}

\subsection{Graph rewiring with virtual nodes }

The use of graph rewiring techniques is inherently linked to computational efficiency considerations. Even though it is possible to design increasingly expressive message  passing architectures, as for instance by enhancing aggregation and update functions to enable information propagation across distant regions of a graph within a limited number of layers, as exemplified by attention-based models, such as Graph Attention Networks (GATs~\cite{gat}),  such approaches often incur substantial computational overhead. This overhead can significantly limit their scalability and practicality on large-scale graphs.
Moreover, even these advanced GNN architectures can benefit from graph rewiring strategies, suggesting that architectural expressiveness alone does not fully address structural limitations in message-passing. Consequently, in many practical settings, simpler architectures such as GCNs or closely related variants are preferred due to their favorable trade-off between expressiveness and computational efficiency, and because their performance can be effectively enhanced through appropriate graph rewiring techniques.

The earliest approach to mitigating over-squashing through virtual nodes introduced a single, globally connected virtual node~\cite{vn1}. This construction effectively reduces the graph diameter to at most two, enabling rapid information propagation across distant nodes. While such methods yield encouraging results on certain datasets, their effectiveness is limited in large or complex graphs, as a single virtual node must encode global information within a constrained representation space.
To address this limitation, subsequent works have proposed the introduction of multiple virtual nodes~\cite{kdd_vn}. By increasing the number of virtual nodes, these methods improve representational capacity and extend applicability to more challenging datasets. However, this comes at the cost of increased computational complexity, as each additional virtual node introduces new edges that substantially raise the message-passing overhead. Furthermore, these approaches significantly alter the original graph topology, which may not be desirable for all tasks.

\begin{figure}[htbp]
\centering
\subfloat[]{
    \includegraphics[width=0.14\textwidth]{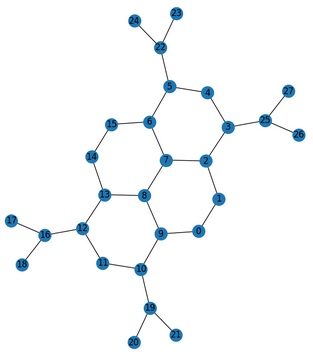}
}
\hfill
\subfloat[]{
    \includegraphics[width=0.14\textwidth]{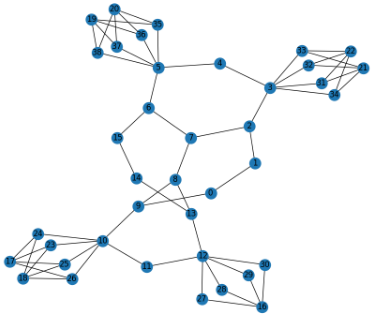}
}
\hfill
\subfloat[]{
    \includegraphics[width=0.14\textwidth]{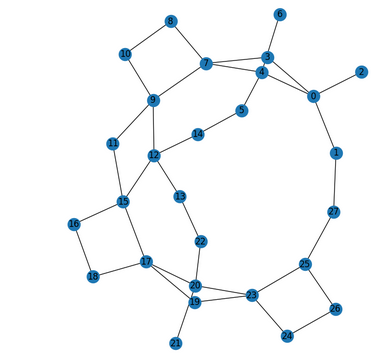}
}
\caption{(a) The original graph;  (b) The gragh generated by  LVN approach; (c) The graph generated using the Expander graph approach}
\label{fig:comp}
\end{figure}

The impact of such topological modifications appears to be dataset-dependent~\cite{survtopo}. Graph rewiring and virtual node techniques have shown strong performance on benchmark graph classification datasets such as MUTAG and PROTEINS, where task-relevant information is largely encoded in the graph structure itself. In contrast, for citation networks such as CORA and CITESEER, node attributes play a dominant role and the underlying topology is less directly correlated with the target labels. In these datasets, aggressive graph rewiring often leads to degraded performance~\cite{local}. This discrepancy suggests that the effectiveness of virtual nodes and rewiring strategies depend on the extent to which task-relevant information is inherently topological, rather than solely relational or feature-driven.
To address this problem, the authors in~\cite{local}  propose the use of multiple Local Virtual Nodes (LVN) rather than a single global virtual node. This approach aims to limit drastic modifications to the graph topology, as depicted in Figure~\ref{fig:comp}, while capturing more localized information and improving the training dynamics of the virtual nodes themselves. While this approach mitigates some of the drawbacks of global virtual nodes, it introduces additional complexity during training, including structural adaptations such as directional virtual nodes, which can substantially increase computational cost.

In this context, PANDA~\cite{pandas}, introduces an approach conceptually similar to virtual nodes by modifying the width of selected nodes, namely, by changing their embedding dimensionality. Intuitively, this mechanism resembles the local virtual nodes introduced in~\cite{local}, as both methods aim to identify central nodes in order to improve local message-passing efficiency by locally increasing the capacity for information storage. However, this approach requires modifying the message-passing functions to accommodate embeddings of multiple dimensionalities. As a consequence, it incurs a higher computational cost than LVN in the undirected setting, while achieving comparable performance.

\section{Related work}

\label{sec:rw}

Several works address the over-smoothing estimation problem~\cite{kdd3}, while other works focus on over-squashing problem~\cite{oversquashing.25}, and analyse and apply different centrality metrics to define the importance of nodes for graph rewiring~\cite{local}.
Some studies explicitly monitor over-smoothing while mitigating over-squashing~\cite{pandas, over_over}, but in most cases these phenomena are treated independently, even when relying on similar theoretical tools, such as spectral analysis~\cite{doc}. In this work, we propose LEED, that takes a first step toward jointly addressing both effects.

\subsection{Over-smoothing estimation}

Most techniques used to evaluate over-smoothing in GNNs rely on quantities equivalent or closely related to the Dirichlet energy~\cite{dirichlet, survey_oversm}, which is computationally efficient and provides interesting results on over-smoothing. 
When this energy becomes low, node embeddings tend to become indistinguishable, resulting in a loss of discriminative information. Conversely, higher Dirichlet energy values correspond to more distinguishable node representations.
However, there is no easy way to describe local over-smoothing effect.
In~\cite{mad}, the authors introduce MAD, an over-smoothing metric grounded in topological properties. While MAD computes cosine similarities between node embeddings on a per-node basis, it relies on a graph-wide filtering mechanism and restricts its analysis to direct neighbors. As a consequence, this formulation is limited in its ability to capture bridge or bottleneck structures, which typically involve interactions beyond immediate neighborhoods. In addition, the use of a global filtering operation reduces the locality of the resulting metric.

To address these limitations, LEED incorporates a corrected two-hop neighborhood aggregation centered around each node. This design enables LEED to capture richer local structural information, including nodes involved in long-range information transfer, while maintaining consistency with global over-smoothing trends. In summary, LEED is more efficient at capturing local properties, while exhibiting graph-level behavior comparable to Dirichlet energy.

\subsection{Finding critical nodes}

A central challenge in these frameworks lies in identifying the nodes that would most benefit from the addition of local virtual nodes, like LVN approaches or adjustments to their embedding width, like PANDA. This problem remains largely unresolved. As a consequence, to address this issue, these frameworks evaluate several classical centrality measures, such as PageRank, degree, betweenness, and closeness, to empirically select the most effective one for a given task. However, this strategy requires testing multiple centrality measures and does not provide a principled criterion for their selection.
In addition, prior theoretical analyses suggest that mitigating over-squashing with graph rewiring is more effectively achieved by strengthening bottleneck or bridge structures rather than emphasizing globally central nodes~\cite{over_over}. This theoretical mismatch highlights the need for alternative criteria that are more directly aligned with the mechanisms underlying over-squashing in GNNs.

In~\cite{VN}, the authors observe that the benefits of virtual nodes are not directly associated with mitigating over-smoothing and may, in some cases, exacerbate it. Nevertheless, over-smoothing is not inherently detrimental and can be advantageous for graph-level tasks, where homogenized node representations facilitate global aggregation. This perspective provides a plausible explanation for the contrasting performance of GNNs augmented with virtual nodes on node-level benchmarks such as CORA and CITESEER, compared to graph-level classification datasets such as MUTAG.
These observations are consistent with the intuition that introducing virtual nodes increases graph connectivity and edge density, which can accelerate the over-smoothing process. While this effect may hinder node-level discrimination, it can be beneficial for tasks that require global structural awareness.

The Local Virtual Node (LVN) framework represents a promising direction by offering flexible mechanisms for localized information aggregation. However, its practical deployment is hindered by the need to tune multiple hyperparameters, including the choice of centrality measure, the number of central nodes considered, and the number of virtual nodes introduced per selected node. In addition, the framework inherits the theoretical limitations associated with classical centrality-based selection and lacks guarantees regarding its impact on training dynamics.
Rather than introducing an entirely new framework with similar limitations, we build upon and refine the LVN paradigm by proposing LEED, a novel metric  specifically designed to improve training dynamics in GNNs. Our approach eliminates the need for dataset-specific centrality selection while directly addressing the structural challenges that arise during message-passing.

\section{LEED}
\label{sec:leed}

LEED (Local Embedding Evolution Distance), 
is a node-level metric designed to characterize over-smoothing in GNNs. 
The basic idea is to quantify the evolution of node embeddings  over the message-passing process to understand how individual node representations change across network layers. In addition, LEED can be used as a unified metric to find critical nodes in graph rewiring strategies aimed at alleviating over-squashing. As a result, LEED offers the key advantage of jointly addressing both over-smoothing and over-squashing problems within a single framework.

\subsection{Design principles}

Ideally, any local over-smoothing evaluation should take the form of a distance measure~\cite{kdd2}, yielding symmetric values in $\mathbb{R}^+$. At the graph level, the notion of symmetry is not directly meaningful; however, at the node level, any measure quantifying the dissimilarity between two nodes should be symmetric. These local distances can then be aggregated to construct a global metric for the entire graph.

Another import aspect to  consider is the trade-off between over-smoothing and over-squashing, as mentioned in Section~\ref{sec:over-over}. These two phenomena are often antagonistic and therefore closely related. As a consequence, adding edges or nodes can alleviate over-squashing by reducing topological bottlenecks in the graph; however, the same modifications may simultaneously exacerbate over-smoothing by increasing connectivity between previously weakly connected regions of the graph. Therefore, by combining solutions designed to address over-smoothing with methods that mitigate over-squashing, it is possible to reduce the risk of trading one problem for the other.

Hence, in this work, we build upon the Dirichlet energy~\cite{dirichlet} and adapt it to our setting by focusing on the identification of critical nodes, since central nodes are as important as bridge nodes in graph rewiring~\cite{over_over}.  We therefore refer to these nodes as {\it critical nodes}, a notion that we define and discuss in detail in the following sections.

The classical Dirichlet energy of an embedding $X^{(k)}$ of a GCN can be calculated by: 

\begin{align}
E(X^{(k)}) 
&= \operatorname{tr}\!\left( {X^{(k)}}^\top \,\tilde{\Delta}\, X^{(k)} \right) \\
&= \frac{1}{2} \sum_{i,j} a_{ij} 
\left\lVert 
\frac{x^{(k)}_i}{\sqrt{1+d_i}} - \frac{x^{(k)}_j}{\sqrt{1+d_j}} 
\right\rVert_2^2 
\end{align}

with $\tilde{\Delta}$ the Laplacian modified operator of the graph, $a_{i,j}$ the adjacency value in $i,j$ and $d_i$ the degree of the adjacency matrix for node $i$.
This formula is computationally efficient and provides interesting results on over-smoothing. However, there is no easy way to describe local over-smoothing effect. We can indeed reduce the embedding $X$ to a specific portion of the signal, but the meaning of the Dirichlet energy is lost in this computation. The over-smoothing phenomenon results from both the GCN architecture and the graph structure. 
As GCN architectures become deeper, node embeddings tend to become increasingly similar.
The same effect is also observed  in  highly connected graphs. The combination of both factors can amplify over-smoothing. 

\begin{figure}[h]
    \centering
    \includegraphics[width=7cm]{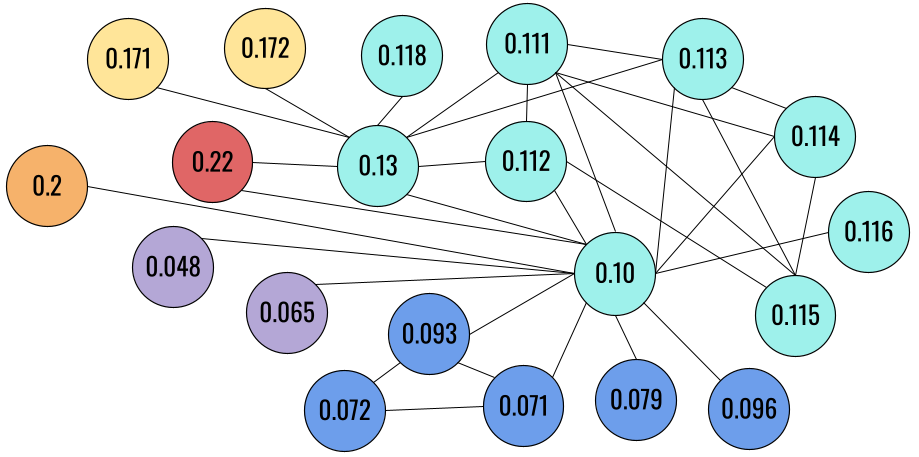}
    \caption{One dimension of node embedding around a clique}
    \label{fig:leed_distance}
\end{figure}

Therefore, in this paper,
rather than viewing over-smoothing as a global optimization concern as is commonly done through regularization terms based on Dirichlet energy, we argue that over-smoothing should not be viewed solely as a message-passing issue, but more fundamentally as a graph connectivity problem: nodes belonging to dense substructures, such as cliques, tend to exhibit embeddings that are significantly more similar to each other than to the rest of the graph.
Figure~\ref{fig:leed_distance} illustrates this effect using one embedding dimension for a clique and its immediate neighborhood after GNN training (two-layer GCN). Nodes within the clique exhibit highly similar embeddings, whereas neighboring nodes can have substantially different representations. Based on our empirical analysis on the CORA dataset, embeddings of nodes within a clique are at least twice as close to each other as to those of their direct neighbors. Moreover, as the number of neighboring nodes decreases, this similarity further increases, reaching up to five times that observed in more diffuse graph regions. In such cases, the absolute value of the Dirichlet energy is not the main concern; rather, the critical issue lies within the clique itself.

Consequently, addressing over-squashing and over-smoothing requires interventions both at the message-passing level and at the level of the graph structure~\cite{hognn}. While many GNN models have been proposed to mitigate these issues, far fewer approaches focus on improving the graph structure for GNN training, and even fewer consider both over-squashing and over-smoothing simultaneously. This gap is partly due to the scarcity of methods capable of performing local evaluations. As with Dirichlet energy, most existing tools provide global solutions, even though the graph structure can and should be treated locally.

\subsection{Local Embedding Evolution Distance}

\begin{figure}[h]
\centering

\subfloat{
    \resizebox{0.3\textwidth}{!}{
        \tikzset{every picture/.style={line width=0.75pt}} 

\begin{tikzpicture}[x=0.75pt,y=0.75pt,yscale=-1,xscale=1]

\draw  [fill={rgb,255:red,10; green,38; blue,180},fill opacity=0.15,  draw=black]  (204,99.5) .. controls (204,93.15) and (209.15,88) .. (215.5,88) .. controls (221.85,88) and (227,93.15) .. (227,99.5) .. controls (227,105.85) and (221.85,111) .. (215.5,111) .. controls (209.15,111) and (204,105.85) .. (204,99.5) -- cycle ;
\draw [fill={rgb,255:red,206; green,16; blue,16},fill opacity=0.15,  draw=black]   (265,106.5) .. controls (265,100.15) and (270.15,95) .. (276.5,95) .. controls (282.85,95) and (288,100.15) .. (288,106.5) .. controls (288,112.85) and (282.85,118) .. (276.5,118) .. controls (270.15,118) and (265,112.85) .. (265,106.5) -- cycle ;
\draw [fill={rgb,255:red,0; green,0; blue,0},fill opacity=0.05,  draw=black]  (227.5,136.5) .. controls (227.5,130.15) and (232.65,125) .. (239,125) .. controls (245.35,125) and (250.5,130.15) .. (250.5,136.5) .. controls (250.5,142.85) and (245.35,148) .. (239,148) .. controls (232.65,148) and (227.5,142.85) .. (227.5,136.5) -- cycle ;
\draw  [fill={rgb,255:red,0; green,0; blue,0},fill opacity=0.05,  draw=black] (310.5,116.5) .. controls (310.5,110.15) and (315.65,105) .. (322,105) .. controls (328.35,105) and (333.5,110.15) .. (333.5,116.5) .. controls (333.5,122.85) and (328.35,128) .. (322,128) .. controls (315.65,128) and (310.5,122.85) .. (310.5,116.5) -- cycle ;
\draw  [fill={rgb,255:red,0; green,0; blue,0},fill opacity=0.05,  draw=black] (222,58.5) .. controls (222,52.15) and (227.15,47) .. (233.5,47) .. controls (239.85,47) and (245,52.15) .. (245,58.5) .. controls (245,64.85) and (239.85,70) .. (233.5,70) .. controls (227.15,70) and (222,64.85) .. (222,58.5) -- cycle ;
\draw  [fill={rgb,255:red,0; green,0; blue,0},fill opacity=0.05,  draw=black] (166,59.5) .. controls (166,53.15) and (171.15,48) .. (177.5,48) .. controls (183.85,48) and (189,53.15) .. (189,59.5) .. controls (189,65.85) and (183.85,71) .. (177.5,71) .. controls (171.15,71) and (166,65.85) .. (166,59.5) -- cycle ;
\draw  [fill={rgb,255:red,0; green,0; blue,0},fill opacity=0.05,  draw=black] (312,71.5) .. controls (312,65.15) and (317.15,60) .. (323.5,60) .. controls (329.85,60) and (335,65.15) .. (335,71.5) .. controls (335,77.85) and (329.85,83) .. (323.5,83) .. controls (317.15,83) and (312,77.85) .. (312,71.5) -- cycle ;
\draw    (250,131) -- (268.5,115) ;
\draw    (310.5,116.5) -- (286.5,112) ;
\draw    (322,105) -- (323.5,83) ;
\draw    (267,101) -- (227,99.5) ;
\draw    (221,89) -- (228,68) ;
\draw    (222,58.5) -- (189,59.5) ;
\draw  [color={rgb, 255:red, 24; green, 22; blue, 170 }  ,draw opacity=1 ][line width=2.25]  (159.5,54.45) .. controls (159.5,31.45) and (253.5,24) .. (257,46) .. controls (260.5,68) and (295.5,92) .. (298.5,121) .. controls (301.5,150) and (207.5,169) .. (202.5,135) .. controls (197.5,101) and (159.5,77.45) .. (159.5,54.45) -- cycle ;
\draw  [color={rgb, 255:red, 204; green, 8; blue, 32 }  ,draw opacity=1 ][line width=1.5]  (206,55) .. controls (212.5,33) and (304,46) .. (330.5,49) .. controls (357,52) and (341.5,102) .. (340.5,128.45) .. controls (339.5,154.9) and (257.5,157) .. (221.5,149) .. controls (185.5,141) and (197.38,124.75) .. (196.88,103.88) .. controls (196.38,83) and (199.5,77) .. (206,55) -- cycle ;
\draw    (346.5,95.77) -- (357,95.77)(346.5,98.77) -- (357,98.77) ;
\draw [shift={(365,97.27)}, rotate = 180] [color={rgb, 255:red, 0; green, 0; blue, 0 }  ][line width=0.75]    (10.93,-3.29) .. controls (6.95,-1.4) and (3.31,-0.3) .. (0,0) .. controls (3.31,0.3) and (6.95,1.4) .. (10.93,3.29)   ;

\draw (273,100) node [anchor=north west][inner sep=0.75pt]   [align=left] [font=\large\bfseries]{i};
\draw (212,92) node [anchor=north west][inner sep=0.75pt]   [align=left] [font=\large\bfseries]{j};
\draw (285,55) node [anchor=north west][inner sep=0.75pt]  [color={rgb, 255:red, 206; green, 16; blue, 16 }  ,opacity=1 ] [align=left] [font=\Large\bfseries]{T};
\draw (295,61) node [anchor=north west][inner sep=0.75pt]  [color={rgb, 255:red, 206; green, 16; blue, 16 }  ,opacity=1 ] [align=left] [font=\Large\bfseries]{{\scriptsize i}};
\draw (164,100) node [anchor=north west][inner sep=0.75pt]  [color={rgb, 255:red, 10; green, 8; blue, 165 }  ,opacity=1 ] [align=left] [font=\Large\bfseries]{T};
\draw (174,106) node [anchor=north west][inner sep=0.75pt]  [color={rgb, 255:red, 10; green, 8; blue, 165 }  ,opacity=1 ] [align=left] [font=\Large\bfseries]{{\scriptsize j}};
\draw (325,86) node [anchor=north west][inner sep=0.75pt]   [align=left][font=\Large\bfseries] {
$l_{i}$};
\draw (200,165.4) node [anchor=north west][inner sep=0.75pt][font=\Large\bfseries]    {$\sum l_{i} =L( X) \sim E( X)$};
\draw (367,90) node [anchor=north west][inner sep=0.75pt]   [align=left] [font=\large\bfseries]{{i}};

\end{tikzpicture}
    }
}\hfill
\subfloat{
    \includegraphics[width=0.16\textwidth]{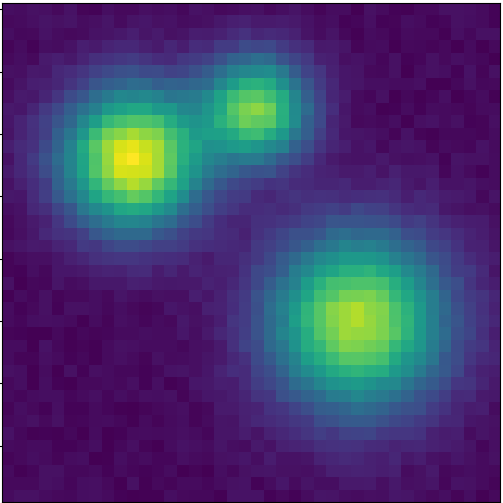}
}

\caption{LEED score and the Heatmap highlighting critical nodes}
\label{fig:fun}
\end{figure}
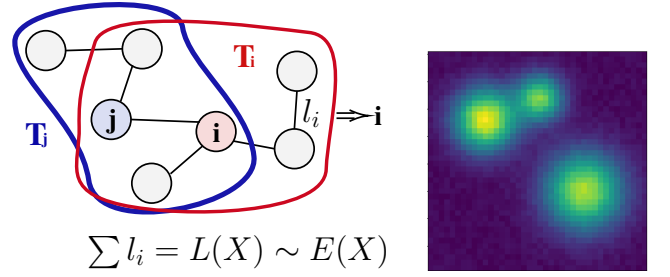

The goal of this work is to propose a tool to enhance GNN performance by addressing both over-smoothing and over-squashing through local graph structure modifications. Therefore, our method enables analysis of the local evolution of node embeddings depending on the underlying graph structure, as illustrated in Figure~\ref{fig:fun}. LEED assigns a local score $l_i$ to each node $i$ based on the 2-hop embedding evolution. Aggregating these scores over all nodes yields a global measure that is comparable to Dirichlet energy, while considering the full set of node-wise scores produces a matrix that allows direct comparison of node importance with respect to the overall over-smoothing, as can be observed in the heatmap.

One way to assess whether a node embedding is close to its neighborhood is to examine how similar its embedding vector is to those of its neighbors. In the Dirichlet energy, this notion is captured by aggregating embedding differences over all neighboring node pairs. In contrast, LEED focuses on local proximity by considering only the minimum distance between a node and its neighbors, rather than summing distances across all edges. Furthermore, instead of directly comparing node embeddings, we apply a transformation prior to distance computation. The Dirichlet distance can be interpreted through the lens of a Markov diffusion kernel~\cite{markov}. In that work, the authors define the Markov diffusion distance as:
\begin{equation}
     d_{i,j}(K)= ||x_i(K) - x_j(K)||_2^2
\end{equation}
With $K$ a time parameter, which can be seen as analogous to the GNN layer index $k$, although it is not equivalent.
That equation leads to a formulation closely related to Dirichlet energy regularization. This connection highlights the close relationship between Markov processes and GCNs, demonstrating that the two are inherently linked. Following this perspective, we propose a Markov evolution process designed to mimic the embedding dynamics of GCNs. Moreover, rather than considering all neighboring pairs as in Dirichlet energy, we retain only the smallest local distance by applying a minimum operator. While the overall computational cost remains comparable (see in \ref{sec:supp} for details), the algebraic formulation is not straightforward. Nevertheless, using a minimum instead of a sum is more appropriate for local distance evaluation: aggregation is meaningful at the graph scale, whereas the minimum provides a sharper characterization of local embedding variations. We define the distance as:
\begin{equation}
    \ell(x^{(k)}_i) = \mathop{\min}\limits_{j \in \mathcal{N}_i} \|T_m(x^{(k)}_i) - T_m(x^{(k)}_j)\|_2^2
\end{equation}

where $T_m$ is a mean evolution function, depending of $m$-hops, weighted by the adjacency matrix where we included a correction for the self-loop: 

 \begin{equation}
      T_1(x^{(k)}_i)=\frac{\mathop{\sum}\limits_{p\in \mathcal{N}_i}a_{ip}x^{(k)}_p+x_i^{(k)}}{\mathop{\sum}\limits_{p\in \mathcal{N}_i} a_{ip}+1}
 \end{equation}

and the idea is to get close to such equation for 2-hops: 

\begin{equation}
    T_2(x_i)=\frac{\mathop{\sum}\limits_{p\in \mathcal{N}_i} a_{i,p}T_1(x_p)+ 2T_1(x_i)}{\mathop{\sum}\limits_{p\in \mathcal{N}_i} a_{i,p}+2}
\end{equation}

However, since node embeddings are not always available, as in the case of the COLLAB dataset, we modify $T_2$ to depend solely on node degree. In this setting, we obtain Equation~\ref{eq:t2}.

\begin{equation}
\label{eq:t2}
    T_2(x^{(k)}_i)=\frac{\mathop{\sum}\limits_{p}a_{ip}\left(\frac{\mathop{\sum}\limits_q a_{pq}x^{(k)}_q.2.(\delta_{i,q}+1)}{\mathop{\sum}\limits_q a_{pq}+2}\right)}{\mathop{\sum}\limits_p a_{ip}+1}
\end{equation}

$\delta_{i,j}$ denotes the Kronecker symbol, which is used to assign more weight to self-loops during the process. In practice, we do not consider more than two-hop neighborhoods. Following~\cite{markov}, we extend this formulation to the entire graph.
by summing over all node embeddings, resulting in:

\begin{equation}
    L(X^{(k)})= \mathop{\sum}\limits_{i=1}^{n} \ell(x^{(k)}_i)
\end{equation}

At the graph level, the Dirichlet energy $E(X^{(k)})$
 exhibits desirable properties that we aim to preserve. In particular, we can derive lower and upper bounds of $E(X^{(k)})$  
 based on the representations at the previous layer:
\begin{equation}
    0 \;\leq\; E\!\left(X^{(k)}\right) \;\leq\; s^{(k)}_{\max} \, E\!\left(X^{(k-1)}\right)    ,
\end{equation}
with $s^{(k)}_{max}$ the squares of maximum singular values of $W^{(k)}$ 
In the original papers, this constraint is used to define two hyperparameters $c_{min}$ and $c_{max}$ to ensure that the Dirichlet energy remains within a suitable range during training:

\begin{equation}
    c_{\min} \, E\!\left(X^{(k-1)}\right) \;\leq\; E\!\left(X^{(k)}\right) \;\leq\; c_{\max} \, E\!\left(X^{(0)}\right)
\end{equation}

Since we aim to preserve this property, we seek either to derive our own bounds or to find a link between $E(X^{(k)})$ and $L(X^{(k)})$. This approach leads, in practice, to the following result:
\begin{equation}
L(X^{(k)})
\le
\left(
\max_{i \in N} \frac{2}{d_i}
\right)
C_{\widehat{T}}
\, E(X^{(k)}),
\label{eq:1}
\end{equation}
with $C_{\widehat{T}}$ a value depending of $T$ and $X$.

\textbf{Proof.} There are two steps in this proof.\\
Let $G=(V,E)$ be an undirected graph with $|V|=n$, adjacency matrix
$A=(a_{ij})_{1\le i,j\le n}$, and degrees
\[
d_i := \sum_{j=1}^n a_{ij}
\]
Let $x_i^{(k)}\in\mathbb{R}^d$ be the feature vector of node $i$ at layer $k$ and
\[
X^{(k)} := (x_1^{(k)},\dots,x_n^{(k)}) \in (\mathbb{R}^d)^n
\]
Step 1: Comparison with unnormalized dirichlet energy.
For each fixed $i$, $T_2(x_i^{(k)})$ is a finite linear combination of the vectors
$\{x_q^{(k)}\}_{q=1}^n$ with coefficients depending only on $A$.
Hence, there exists a matrix $V=(V_{iq})\in\mathbb{R}^{n\times n}$ such that
\[
T_2(X^{(k)}) = V X^{(k)}
\]

Since $\mathcal{N}_i$ is finite and nonempty for all i,
For
\[
S_i:=\{ ||T_2(x_i)-T_2(x_j)||_2^2, j \in \mathcal{N}_i\}
\]
We have: 
\[
\min S_i \leq \overline{S_i}=\frac{1}{|\mathcal{N}_i|}\sum_{j\in\mathcal{N}_i} a_j ||T_2(x_i)-T_2(x_j)||_2^2
\]

Applying this inequality for a symmetric binary matrix, and summing over nodes ($d_i=|\mathcal{N}_i|$) we get:

\[
L(X^{(k)})
\le
\sum_{i=1}^n
\frac{1}{d_i}
\sum_{j=1}^n a_{ij}
\left\|
T_2(x_i^{(k)}) - T_2(x_j^{(k)})
\right\|_2^2
\tag{$\star$}
\]

Define the (unnormalized) Dirichlet energy of a signal $Z=(z_1,\dots,z_n)$ by
\[
\mathcal{E}(Z)
:=
\frac{1}{2}
\sum_{i,j=1}^n
a_{ij}
\| z_i - z_j \|_2^2 
\]

From $(\star)$ we obtain
\[
L(X^{(k)})
\le
\left(
\max_{i \in N} \frac{2}{d_i}
\right)
\mathcal{E}(T_2(X^{(k)}))
\]

This step explain why the profile of the function has good reason to be similar to the Dirichlet energy function. However, we are using a normalized Dirichlet energy, and that normalization explain the difference of range in the values proposed:  

\begin{align*}
E(X^{(k)}) :=\;
& \frac{1}{2}
\sum_{i,j=1}^n a_{ij}
\left\|
\frac{x_i^{(k)}}{\sqrt{1+d_i}}
-
\frac{x_j^{(k)}}{\sqrt{1+d_j}}
\right\|_2^2 \\
=\;
& \mathcal{E}(\tilde D^{-1/2} X^{(k)})
\end{align*}

Step 2: Extension using conjugated operator.
Define the linear operator
\[
\widehat T
:=
\tilde D^{-1/2} \, V \, \tilde D^{1/2}
\]
Then
\[
\mathcal{E}(T_2(X^{(k)}))
=
\mathcal{E}(\widehat T \, \tilde D^{-1/2} X^{(k)})
\]

Let

\[
C_{\widehat{T}} := \sup_{Y \notin \ker(\Delta)} \frac{\mathcal{E}(\widehat{T} Y)}{\mathcal{E}(Y)} = \sup_{Y \notin \ker(\Delta)} \frac{\mathrm{Tr}(Y^\top \widehat{T}^\top \Delta \widehat{T} Y)}{\mathrm{Tr}(Y^\top \Delta Y)}
\]

Applying this with $Y=\tilde D^{-1/2} X^{(k)}$ yields
\[
L(X^{(k)})
\le
\left(
\max_{i \in N} \frac{2}{d_i}
\right)
C_{\widehat{T}}
\, E(X^{(k)}),
\]
with for most graphs:
$$ \left(
\max_{i \in N} \frac{2}{d_i}
\right) =2$$
\hfill $\square$

As some nodes (e.g., leaf nodes) are not taken into account in the computation of $T_2$, we cannot obtain similarly infimum.
However, with all simulations we can easily see that in practice $ \exists C$ s.t $C.E(X^{(k)}) \leq L(X^{(k)})$, leading to equivalent measurement for E and L except in special cases.

\begin{figure*}[t]
\centering

\subfloat{
    \includegraphics[width=0.23\textwidth]{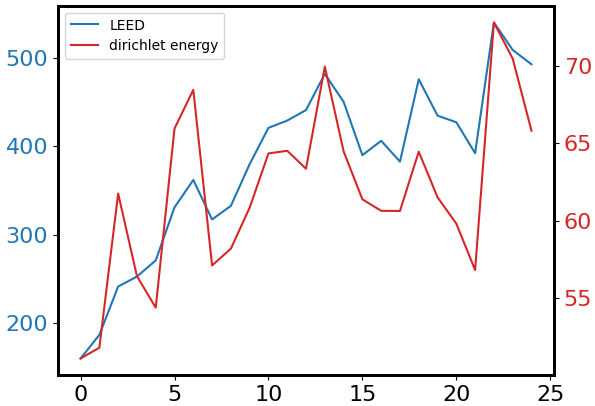}
}\hfill
\subfloat{
    \includegraphics[width=0.23\textwidth]{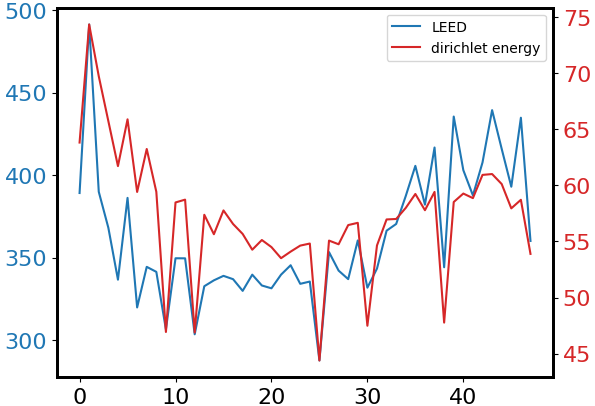}
}\hfill
\subfloat{
    \includegraphics[width=0.23\textwidth]{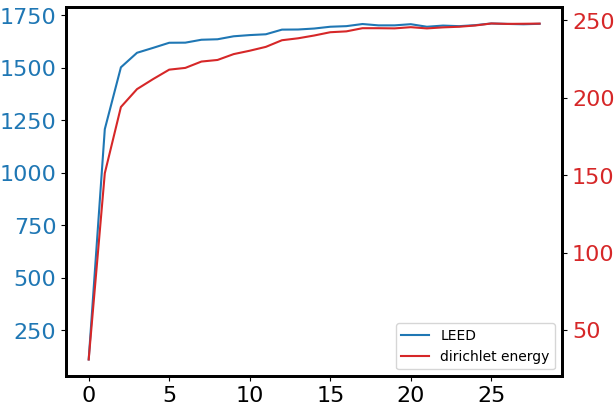}
}\hfill
\subfloat{
    \includegraphics[width=0.23\textwidth]{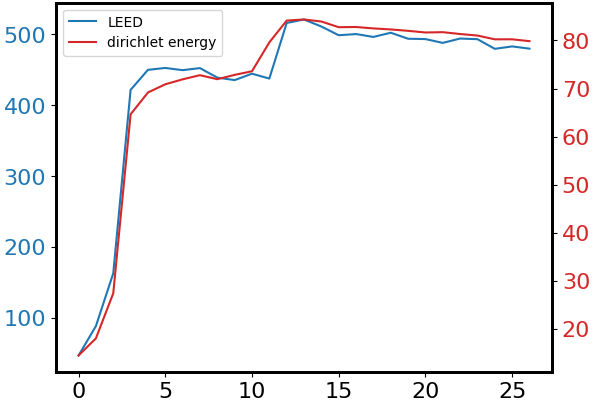}
}

\medskip

\subfloat{
    \includegraphics[width=0.23\textwidth]{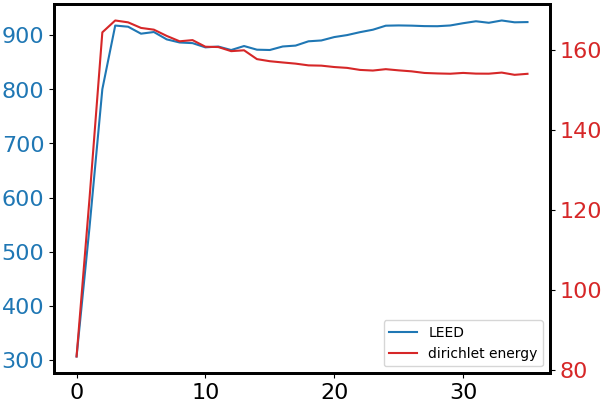}
}\hfill
\subfloat{
    \includegraphics[width=0.23\textwidth]{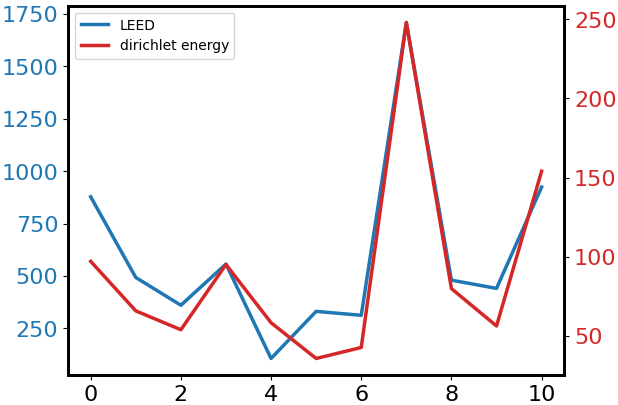}
}\hfill
\subfloat{
    \includegraphics[width=0.23\textwidth]{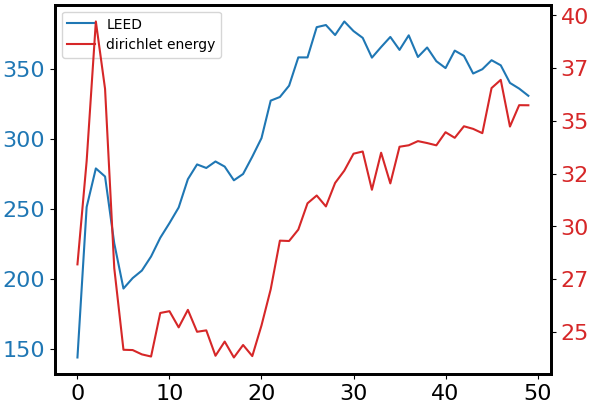}
}\hfill
\subfloat{
    \includegraphics[width=0.23\textwidth]{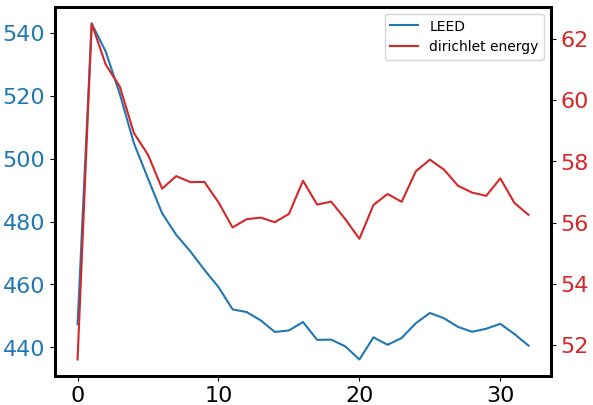}
}

\caption{Comparison with 8 configurations between LEED and Dirichlet energy}
\label{fig:leed_cmp}
\end{figure*}

\begin{figure}[t]
    \centering
    \includegraphics[width=7cm]{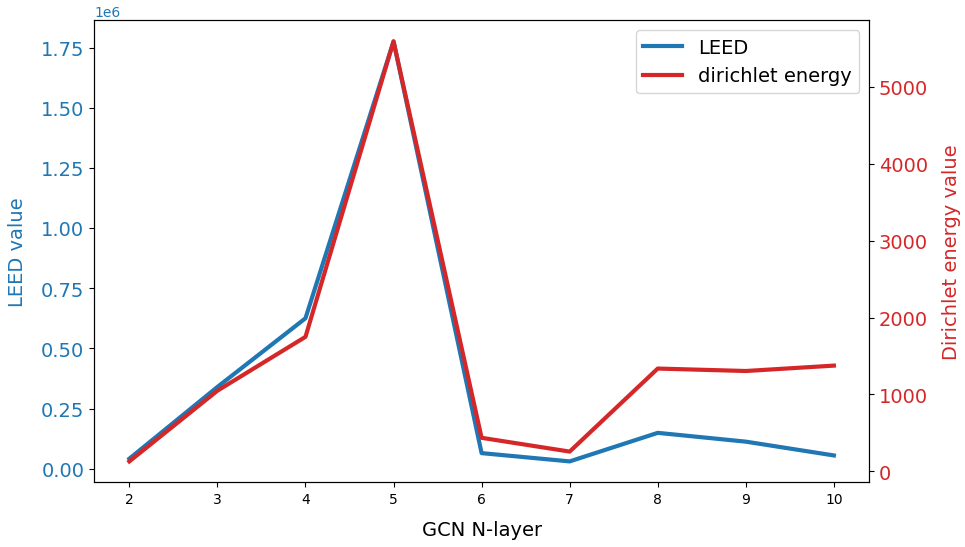}
    \caption{Cora distance for each layer between 2 and 10}
    \label{dirleed}
\end{figure}

\begin{figure}[t]
    \centering
       \includegraphics[width=7cm]{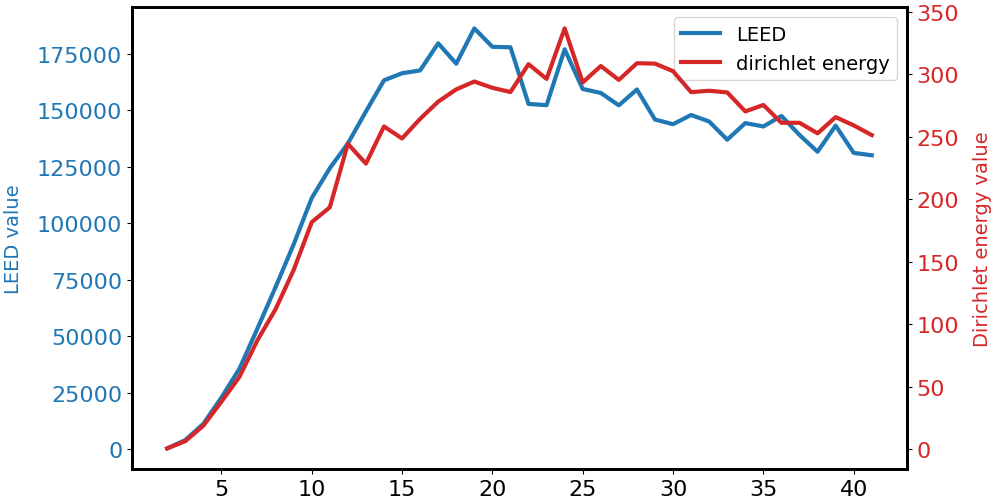}
    \caption{Cora distance for every 10 epoch of training}
    \label{fig:large}
\end{figure}

In practice $C_{\widehat{T}}$ can become very large for large-scale datasets, as illustrated in Figures~\ref{dirleed}~and~\ref{fig:large}. However,  results also suggest that the profile of Dirichlet energy is likely to be similar to that of the LEED distance.
Since the relative evolution of LEED is more informative than its absolute value, we design an experiment to validate this behavior. Specifically, we conduct a series of comparative experiments using eight different adjacency matrices for GNN training, and analyze whether the trajectories of Dirichlet energy and LEED over training epochs exhibit similar trends. The results are presented in Figure~\ref{fig:leed_cmp}. We establish a link between these two functions, which can be exploited to impose similar constraints during training to prevent over-smoothing. We perform additional experiments to study the evolution of this distance as a function of the number of GCN layers, evaluating both LEED and the Dirichlet energy at the final layer. Results are reported in Figure~\ref{dirleed}. As the CORA dataset is not connected, the over-smoothing metrics are not expected to converge to zero.
In addition, the resulting function presents local dependencies, which allows us to assess how individual nodes locally affect its score. Further investigation of these properties is left for future work.

\section{Critical node selection}
\label{sec:cns}

An important advantage of LEED is its ability to identify critical nodes for GNN performance. We define critical nodes as those that play a decisive role in information propagation and representation learning within the graph. This notion is motivated by prior work showing that node importance, often formalized through centrality measures, is instrumental in mitigating over-squashing~\cite{local}, while recent theoretical analyses emphasize the role of bottleneck structures in limiting long-range information flow~\cite{curv}.

To reconcile these perspectives, we design a critical node score that jointly captures centrality-related effects, bottleneck sensitivity, and embedding distance evaluation. Specifically, the proposed score integrates local structural influence with the behavior of node embeddings during training, allowing it to reflect both topological and representational constraints.
To this end, we reuse the local function $T$, previously introduced, and apply it in a comparative setting by incorporating input embedding values as a reference:
Let 
\begin{equation}
    D_i = \bigl\|\, T_2(x_i^{(0)}) - x_i^{(0)} \,\bigr\|_2, \quad i \in [1, n]
\end{equation}

As we do not need to search among each layers anymore, we instead use only input values and initial embeddings. Then, we obtain a set of critical nodes by taking a top-K, which is
denoted by $\tau_k$ the
$(1-k)$-quantile of the set $\{D_i\}_{i=1}^n$, defined as:
\begin{equation}
    \tau_k = \inf \Bigl\{\, t \in \mathbb{R} \;:\; 
\frac{1}{n}\, \#\{\, i \in [1,n] \;|\; D_i \ge t \,\} \le k \Bigr\}
\end{equation}

Hence, the ensemble of critical nodes corresponding to the top $k = \tfrac{X}{100}$ fraction of the $D_i$ is given by
\begin{equation}
    \mathcal{S}_k = \bigl\{\, i \in [1,n] \;\big|\; D_i \ge \tau_k \,\bigr\}
\end{equation}

\subsection{Properties on critical nodes}

\label{sec:res}

Our formulation assumes the availability of node embeddings. However, certain benchmark datasets, such as REDDIT-BINARY and COLLAB, do not provide explicit node features. In such cases, we adopt a standard strategy commonly used in graph rewiring methods~\cite{cayley, pandas} that assigns a constant feature vector $\mathbf{1}_n$ to all nodes. This choice corresponds to initializing each node with an identical scalar embedding. In practice, we restrict the embedding to a single dimension, as scores computed across different embedding dimensions differ only by a multiplicative factor. For undirected graphs with a symmetric binary adjacency matrix, this simplification allows the proposed measure to be expressed in closed form, yielding the following output:
\begin{equation}
  D_i=\frac{|d_i-2|}{d_i+2}   ,
\end{equation}

except when $ d_p<< d_i, \forall p\in \mathcal{N}_i$. This simple formula yields a value    in $[0,1]$, that increases monotonically with $d_i$, the degree of node $i$. When  seeking for solutions involving more complex embedding such as one-hot encoding of dimension $d$, present in MUTAG, ENZYMES datasets, we obtain: 

$$\mathbb{E}[D_i^2] = \alpha_i^2 + \left( 1 - \frac{1}{d} \right) \sum_{q \neq i} \beta_q^2 + \frac{2\alpha_i}{d} \sum_{q \neq i} \beta_q + \frac{1}{d} \left( \sum_{q \neq i} \beta_q \right)^2 $$

with $$\alpha_i=\frac{4}{d_i+1}\sum_{p\in \mathcal{N}_i} \frac{1}{d_p+1}-1$$
and $$
\beta_q = \frac{2}{d_i+1} \sum_{p \in \mathcal{N}(i)} \frac{\mathds{1}_{\{q \in \mathcal{N}(p) \setminus \{i\}\}}}{d_p + 2}$$

This suggests that the score for $D_i$ is high when node $i$ has a limited number of neighbors $p_i$, but each of these neighbors has several neighbors $q_p$ ($\alpha$ parameter). Additionally, the value of $\beta_q$ tends to be higher when there is no connection between any pair of $p$-nodes. 
As a result, the optimal score for node
$i$ arises when it serves to bridge distinct components of the graph. Consequently, the distance measure places greater emphasis on bridging structures, similar to curvature-based methods~\cite{doc}. Hence, LEED inherits the well-established advantages for training GNNs.
However, this solution applies only for a normal random distribution of the one-hot embedding. If the distribution shifts, a new parameter emerges, representing the distance between the embedding and its direct neighbors. This results in a different type of centrality measure, one that is more closely related to information importance than to topological importance~\cite{kdd2}.

\section{Experimental implementation} 
\label{subsection:experiment}

In our experimental evaluation, we compare LEED to a standalone GCN model and other three graph rewiring approaches: Cayley graph~\cite{cayley}, LVN~\cite{local}, and  PANDA~\cite{pandas}. For their implementation, we rely on open-source code that are publicly available  in github. For benchmarking, we  evaluate their performance with six widely adopted datasets from the TUDataset collection~\cite{tudata}: MUTAG, ENZYMES, PROTEINS, REDDIT-BINARY, IMDB-BINARY, and COLLAB. These datasets are commonly used to evaluate a model’s ability to capture long-range interactions in graph-level tasks.
We split the datasets as 60\% for the training set, 20\% for the validation set and 20\% for the test set.

Results for the standalone GCN model are taken directly from the Cayley graph paper, with the exception of PROTEINS, where improved performance is reported under the LVN framework and is not reproduced in our experiments. All other results are reproduced using dataset-specific hyperparameter configurations.
For Cayley graph rewiring, we use the results reported in the original paper as baselines, as the full set of hyperparameters is not provided. Nevertheless, we verified that comparable performance can be achieved across all datasets under similar experimental conditions.
For LVN, we extensively evaluated the framework using the hyperparameters provided by the authors, conducting 20 independent runs for reproducibility. Given the lower variance reported in the original results, we retain their reported scores.
Finally, for PANDA, we employ the original framework and use the same hyperparameter search space as described by the authors, with 30 runs for MUTAG. For the remaining datasets, the search space is reduced based on the best configurations reported in the original work. We observe improved performance on two datasets and report, for each case, the best result between our experiments and those from the original paper.

To evaluate the effectiveness of LEED in selecting critical nodes for graph rewiring, we apply it to LVN~\cite{local} and PANDA~\cite{pandas}. In the original methods, several centrality metrics were used to determine central nodes. Our objective here is to replace all of these metrics with our proposed local distance measure. We keep all the different hyperparameters of the original frameworks. For example, both methods employ a top-K strategy to obtain the set of central nodes, and a number called `worker' to create multiple virtual nodes for each central node selected. Similarly, PANDA uses an expansion parameter for the width of embeddings.
We do not include a comparison with a directed LVN in the main evaluation, as it does not align with the mathematical framework of LEED. However, we conducted additional experiments to explore this setting, which can be found in \ref{sec:supp}, we leave further exploration of this possibility for future work. Whenever the code was available, we ran the experiments ourselves.
For all our experiments, we use a server with 2 GPU NVIDIA RTX 6000 Ada Generation (49 Go).

\section{Results}
\label{sec:results}

\begin{table*}[t]
\centering
\caption{Classification accuracy (\%) on benchmark datasets, with \textcolor{red}{first}, \textcolor{blue}{second} and \textcolor{violet}{third} best scores}
\resizebox{\textwidth}{!}{
    \begin{tabular}{c c c c c c c c}
        \hline
        \textbf{Model} & \textbf{MUTAG} & \textbf{ENZYMES} & \textbf{PROTEINS} & \textbf{REDDIT-Binary} & \textbf{IMDB-Binary} & \textbf{COLLAB} & \textbf{Avg. Rank} \\
        \hline
        GCN & 74.750$\pm$4.030 & 29.083$\pm$2.363 & 71.410$\pm$1.077 & 77.735$\pm$1.586 & 60.500$\pm$2.729 & \textcolor{violet}{70.490$\pm$1.628} & 5.1 \\
        \hline
        Cayley~\cite{cayley} & \textcolor{violet}{83.750$\pm$3.597} & 31.000$\pm$2.397 & 73.036$\pm$1.291 & 67.050$\pm$1.483 & 56.200$\pm$1.825 & 69.630$\pm$0.730 & 4.8 \\
        \hline

        LVN~\cite{local} & 82.333$\pm$2.149 & \textcolor{violet}{31.367$\pm$1.376} & 73.189$\pm$0.882 & \textcolor{blue}{83.440$\pm$0.775} & \textcolor{blue}{66.620$\pm$1.513} & \textcolor{blue}{71.520$\pm$0.661} & 3 \\
        \hline
        PANDA~\cite{pandas} & \textcolor{blue}{86.068$\pm$2.198} & \textcolor{blue}{31.550$\pm$1.230}& \textcolor{blue}{75.647$\pm$1.076}& \textcolor{violet}{80.690$\pm$0.721} & 63.760$\pm$1.012 & 68.433$\pm$1.012 & 3 \\
        \hline
        \hline
        PANDA-LEED & \textcolor{red}{86.838$\pm$1.653} &  \textcolor{red}{33.182$\pm$1.921} & \textcolor{red}{75.781$\pm$1.413}& 80.488$\pm$0.946 & \textcolor{violet}{64.700$\pm$1.748} & 67.785$\pm$1.001 & 2.6 \\
        \hline
        LVN-LEED & 83.333$\pm$2.325 & 31.094$\pm$1.748 & \textcolor{violet}{74.775$\pm$1.066} & \textcolor{red}{84.330$\pm$0.979} & \textcolor{red}{68.060$\pm$1.533} & \textcolor{red}{72.684$\pm$0.567} & 2.3 \\
        \hline

    \end{tabular}
}
\label{tab:results}
\end{table*}

Table~\ref{tab:results} summarizes our results. It is clear that LVN with LEED performs globally better than any other model. We also obtain better results with LVN-LEED on almost all datasets compared to classical LVN.
In addition, PANDA-LEED achieves strong performance on the first three datasets. For graph-structured datasets with informative node embeddings, such as MUTAG, ENZYMES, and PROTEINS, the results indicate that LEED reliably identifies critical nodes, which can subsequently be leveraged to construct meaningful virtual nodes or enlarged node representations.
In contrast, 
for datasets without intrinsic node embeddings, LEED also attains competitive performance, yielding good results in most cases. The only exception is COLLAB within the PANDA framework. However as PANDA alone performs even worse than classical GCN on this dataset, it is not surprising to observe a performance degradation, even if not statistically significant, when attempting to further refine PANDA. These findings suggest that even in the absence of node features, LEED provides a reliable estimator of node importance, performing comparably to, and in many cases better than, existing approaches.
Finally, it is worth mentioning that training increasingly large models with numerous hyperparameters becomes progressively more expensive. Hence, achieving better or even comparable performance with fewer parameters is a highly encouraging result. Using only LEED instead of testing several classical distance metrics to find critical nodes is definitely more efficient. 

\subsection{Supplementary tests}
\label{sec:supp}
In Table~\ref{tab:complex} we show classical metrics and Dirichlet energy complexity for an unweighted sparse graph to compare with LEED and in particular $T_2$ function. With $m=|E|$, $n=|V|$, $l$ the dimension of embeddings and $\Bar{d}$ the mean degree of nodes. It shows that for sparse graphs, where $\Bar{d}$ is low, LEED will be approximately as fast as Dirichlet energy.

\begin{table}[h!]
    \centering
    \caption{Complexity centrality metrics vs LEED}
\begin{tabular}{|c|c|}
    \hline
    PageRank (k iter) & $\mathcal{O}(k.m)$   \\
    Closeness & $\mathcal{O}(n.m)$ \\
    Betweenness & $\mathcal{O}(n.m)$ \\
    Dirichlet &  $\mathcal{O}(l.(m+n))$ \\
    LEED & $\mathcal{O}(l.n.\Bar{d})$\\
    \hline
\end{tabular}
\label{tab:complex}
\end{table}

The framework LVN is also producing a directed graph GCN model. Since we developed LEED with a symmetric adjacency matrix in mind, and most of the literature and available mathematical tools assume this structure, we did not investigate this aspect in depth. However, we conducted several tests with directed LVN for comparison, using its original values. As shown in Table~\ref{tab:directed}, we obtain results comparable to prior work on MUTAG and PROTEINS, while significantly improving the previous best performance on ENZYMES.

\begin{table}[h!]
    \centering
    \caption{Directed LVN comparison\\
    * best among:\{degree, betweenness, closeness, pagerank\}}
    \begin{tabular}{|c|c|c|c|} \hline
         & MUTAG & ENZYMES & PROTEINS \\ \hline
         original metrics$^{*}$ & \textcolor{red}{84.778$\pm$2.902}  & 31.400$\pm$2.095 &74.360$\pm$1.105 \\
         \hline
        with LEED &  83.444$\pm$2.694 &  \textcolor{red}{37.773$\pm$1.669} & \textcolor{red}{74.793$\pm$1.138}\\
        \hline
    \end{tabular}
    \label{tab:directed}
\end{table}

\subsection{Critical nodes on over-smoothing} 
\begin{figure}[t]
    \centering
    \includegraphics[width=8.5cm]{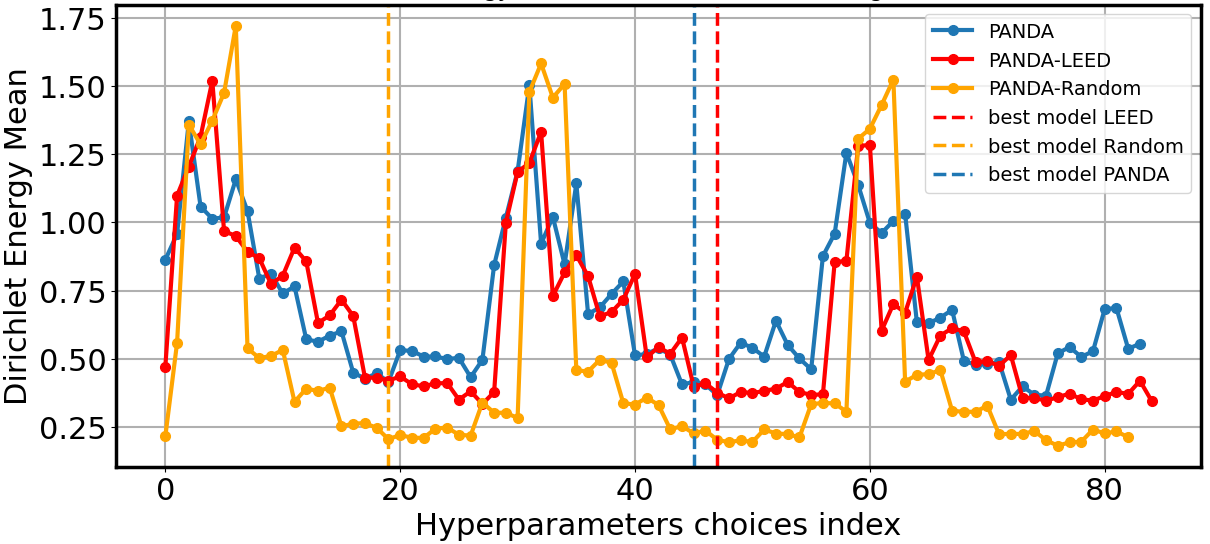}
    \caption{Dirichlet energy over sets of hyperparameters for LEED and other metrics in MUTAG dataset}
    \label{fig:overs}
\end{figure}

As previously discussed, over-smoothing is no longer universally regarded as either beneficial or detrimental; its impact is highly task-dependent, and different forms or degrees of over-smoothing can either degrade or improve performance~\cite{VN}. Architectural modifications such as introducing virtual nodes or increasing the embedding width of selected nodes typically exacerbate over-smoothing. While prior works has largely focused on mitigating this effect, over-smoothing can also be leveraged to facilitate the propagation of important information and, in some cases, improve the model performance.
To assess the impact of critical node selection on over-smoothing, we compare PANDA with LEED and random selection across varying hyperparameters.
Figure~\ref{fig:overs} illustrates that optimal hyperparameter configurations are not necessarily associated with a reduction of over-smoothing. We observe that randomly selecting central nodes in the PANDA framework tends to increase the risk of over-smoothing, whereas our critical node selection achieves performance comparable to that obtained with the best-performing classical centrality measures.

\section{Conclusion}
\label{sec:conc}

In this paper, we introduced LEED, a local metric for quantifying over-smoothing that preserves a global behavior comparable to Dirichlet energy while enabling node-level analysis. We demonstrated that this locality can be effectively exploited within virtual and expanded nodes mechanisms to mitigate over-squashing. 
Our results suggest that over-smoothing and over-squashing should be addressed jointly, as they are intrinsically linked through the dynamics of message-passing in GNNs and the graph topology.

Empirically, LEED  outperforms most classical centrality measures when used to guide both PANDA and LVN construction across a wide range of graph-level tasks, indicating strong consistency in the LEED metric.
Our experimental evaluation is restricted to the GCN architecture, whereas many rewiring and virtual node methods have also been explored in conjunction with alternative architectures such as GIN. Extending the analysis to these models constitutes a natural direction for future work. Moreover, while we focus on the application of LEED within virtual node frameworks, the broader implications of local over-smoothing analysis remain largely unexplored. Leveraging LEED to design novel, adaptive graph rewiring strategies represents a promising avenue for further investigation.

\bibliographystyle{ieeetr}
\bibliography{biblio2}

@article{oversquashing.25,
title = {Over-squashing in Graph Neural Networks: A comprehensive survey},
journal = {Neurocomputing},
volume = {642},
pages = {130389},
year = {2025},
author = {S. Akansha}
}

@article{GCN,
  title={Semi-supervised classification with graph convolutional networks},
  author={Thomas N. Kipf and Max Welling},
  journal={arXiv preprint arXiv:1609.02907},
  year={2016}
}

@article{expressive,
  title={Graph neural networks exponentially lose expressive power for node classification},
  author={Oono, Kenta and Suzuki, Taiji},
  journal={arXiv preprint arXiv:1905.10947},
  year={2019}
}

@article{survey_oversm,
  title={A survey on oversmoothing in graph neural networks},
  author={Rusch T Konstantin and Bronstein Michael M and Mishra Siddhartha},
  journal={arXiv preprint arXiv:2303.10993},
  year={2023}
}

@article{dirichlet,
  title={Dirichlet energy constrained learning for deep graph neural networks},
  author={Zhou Kaixiong and Huang Xiao and Zha Daochen and Chen Rui and Li Li and Choi Soo-Hyun and Hu Xia},
  journal={Advances in neural information processing systems},
  volume={34},
  pages={21834--21846},
  year={2021}
}

@inproceedings{mad,
    title={measuring and relieving the over-smoothing problem for graph neural networks from the topological view},
    author={Deli Chen and Yankai Lin and Wei Li and Peng Li and Jie Zhou and Xu Sun},
    year={2020},
    booktitle = {AAAI'20}
}

@article{curvature1,
  title={Understanding over-squashing and bottlenecks on graphs via curvature},
  author={Topping Jake and Di Giovanni Francesco and Chamberlain Benjamin Paul and Dong Xiaowen and Bronstein Michael M},
  journal={arXiv preprint arXiv:2111.14522},
  year={2021}
}

@inproceedings{SGCN,
  title={Simplifying graph convolutional networks},
  author={Wu Felix and Souza Amauri and Zhang Tianyi and Fifty Christopher and Yu Tao and Weinberger Kilian},
  booktitle={ICML'19},
  year={2019}
}

@inproceedings{VN,
    title= {understanding virtual nodes: oversquashing and node heterogeneity},
    author={Joshua Southern and Francesco Di Giovanni and Michael Bronstein and Johannes F. Lutzeyer},
    booktitle = {ICLR'25},
    year= {2025}}

@article{cayley,
    title = {Cayley Graph Propagation},
    journal = {ArXive},
    author= {JJ Wilson and Maya Bechler-Speicher and Petar Veličković},
    year= {2025}}

@article{local,
    title={Local Virtual Nodes for Alleviating
    Over-Squashing in Graph Neural Networks},
    journal = {ArXive},
    author= {Tuğrul Hasan Karabulut and İnci M. Baytaş},
    year = {2025}
}

@inproceedings{vn1,
    author = {Cai Chen and Hy Truong Son and Yu Rose and Wang Yusu},
    title = {On the connection between {MPNN} and graph transformer},
    year = {2023},
    booktitle = {ICML'23},
}

@inproceedings{computer_v,
  author={Krzywda Maciej and Łukasik Szymon and Gandomi Amir H.},
  booktitle={International Joint Conference on Neural Networks (IJCNN)}, 
  title={Graph Neural Networks in Computer Vision - Architectures, Datasets and Common Approaches}, 
  year={2022}
  }

@inproceedings{gat,
    title={Graph Attention Networks},
    year={2018},
    author={Petar Velickovic and Guillem Cucurull and Arantxa Casanova and Adriana Romero and Pietro Lio and Yoshua Bengio},
    booktitle = {ICML'18}

}

@inproceedings{over_over,
    author = {Giraldo Jhony H. and Skianis Konstantinos and Bouwmans Thierry and Malliaros Fragkiskos D.},
    title = {On the Trade-off between Over-smoothing and Over-squashing in Deep Graph Neural Networks},
    year = {2023},
    booktitle = {ACM International Conference on Information and Knowledge Management (ICKM}
}

@article{doc,
    title={Rewiring Techniques to Mitigate Oversquashing
    and Oversmoothing in {GNN}s: A Survey},
    author={Hugo Attali and Davide Buscaldi and Nathalie Pernelle},
    year={2024},
    journal={arXiv}
}

@inproceedings{pandas,
    title = {{PANDA}: Expanded Width-Aware Message Passing Beyond Rewiring},
    author = {Jeongwhan Choi and Sumin Park and Hyowon Wi and Sung-Bae Cho and Noseong Park},
    year = {2024},
    booktitle = {ICML'24},
}

@inproceedings{kdd_vn,
    title={Virtual Node Tuning for Few-shot Node Classification},
    booktitle = {ACM SIGKDD '23},
    author={Zhen Tan and Ruocheng Guo and Kaize Ding and Huan Liu},
    year={2023}
}

@inproceedings{markov,
    title = {Simple spectral graph convolution},
    author = {Hao Zhu and Piotr Koniusz},
    year = {2021},
    booktitle ={ICLR'21}
}

@inproceedings{kdd2,
    author = {Chen Rongqin and Li Yan and Wu Dan and Mo Fan and Zhang Shenghui and Ip Pak Lon and Iam Hoi Cheong and Li Ye and U Leong Hou},
    title = {Enhanced Subgraph Learning in 2-{FWL} {GNN}s via Local Connectivity, Spectral, and Distance Encodings},
    year = {2025},
    booktitle = {ACM SIGKDD '25}
}

@inproceedings{curv,
    title={The Effectiveness of Curvature-Based Rewiring and the Role of Hyperparameters in {GNN}s Revisited},
    author = {Floriano Tori and Vincent Holst and Vincent Ginis},
    booktitle = {ICLR'25},
    year={2025},
}

@inproceedings{network,
    author = {Almasan Paul and Su\'{a}rez-Varela Jos\'{e} and Lutu Andra and Cabellos-Aparicio Albert and Barlet-Ros Pere},
    title = {Enhancing {5G} Radio Planning with Graph Representations and Deep Learning},
    year = {2023},
    booktitle = {5G-MeMU '23}
}

@inproceedings{express,
    author = {Chen Kaixuan and Liu Shunyu and Zhu Tongtian and Qiao Ji and Su Yun and Tian Yingjie and Zheng Tongya and Zhang Haofei and Feng Zunlei and Ye Jingwen and Song Mingli},
    title = {Improving Expressivity of {GNN}s with Subgraph-specific Factor Embedded Normalization},
    year = {2023},
    booktitle = {ACM SIGKDD '23}
}

@inproceedings{kdd3,
    author = {Zhao Weichen and Wang Chenguang and Wang Xinyan and Han Congying and Guo Tiande and Yu Tianshu},
    title = {Understanding Oversmoothing in Diffusion-Based {GNN}s From the Perspective of Operator Semigroup Theory},
    year = {2025},
    booktitle = {ACM SIGKDD '25}
}

@inproceedings{log2,
  title={Mitigating over-smoothing and over-squashing using augmentations of forman-ricci curvature},
  author={Fesser Lukas and Weber Melanie},
  booktitle={Learning on Graphs Conference},
  year={2024}
}

@inproceedings{tudata,
  title = {Tudataset: A collection of benchmark datasets for learning with graphs},
  author = {Morris Christopher and Kriege Nils M and Bause Franka and Kersting Kristian and Mutzel Petra and Neumann Marion},
  booktitle = {ICML'20},
  year={2020}
}

@ARTICLE{journal1,
  author={Chen Tianlong and Zhou Kaixiong and Duan Keyu and Zheng Wenqing and Wang Peihao and Hu Xia and Wang Zhangyang},
  journal={IEEE Transactions on Pattern Analysis and Machine Intelligence}, 
  title={Bag of Tricks for Training Deeper Graph Neural Networks: A Comprehensive Benchmark Study}, 
  year={2023},
  volume={45},
  number={3},
  pages={2769-2781}}

@ARTICLE{dyna2,
  author={Dong Wei and Yan Dawei and Wang Peng},
  journal={IEEE Transactions on Pattern Analysis and Machine Intelligence}, 
  title={Self-Supervised Node Representation Learning via Node-to-Neighbourhood Alignment}, 
  year={2024},
  volume={46},
  number={6},
  pages={4218-4233}}

@ARTICLE{multivn,
  author={Zhang Yuelin and Cen Jiacheng and Han Jiaqi and Huang Wenbing},
  journal={IEEE Transactions on Pattern Analysis and Machine Intelligence}, 
  title={Fast and Distributed Equivariant Graph Neural Networks by Virtual Node Learning}, 
  year={2026},
  pages={1-15}}

@ARTICLE{hognn,
  author={Maciej Besta et al.},
  journal={IEEE Transactions on Pattern Analysis and Machine Intelligence}, 
  title={Demystifying Higher-Order Graph Neural Networks}, 
  year={2026},
  volume={48},
  number={3},
  pages={2544-2565},
}

@ARTICLE{gstruct,
  author={Peng Liang and Hu Rongyao and Kong Fei and Gan Jiangzhang and Mo Yujie and Shi Xiaoshuang and Zhu Xiaofeng},
  journal={IEEE Transactions on Neural Networks and Learning Systems}, 
  title={Reverse Graph Learning for Graph Neural Network}, 
  year={2024},
  volume={35},
  number={4},
  pages={4530-4541}}

@ARTICLE{survtopo,
  author={Pham Phu and Bui Quang-Thinh and Thanh Nguyen Ngoc and Kozma Robert and Yu Philip S. and Vo Bay},
  journal={IEEE Transactions on Neural Networks and Learning Systems}, 
  title={Topological Data Analysis in Graph Neural Networks: Surveys and Perspectives}, 
  year={2025},
  volume={36},
  number={6},
  pages={9758-9776}}

@inproceedings{gradrew,
author = {Jiang Zhimeng and Liu Zirui and Han Xiaotian and Feng, Qizhang and Jin Hongye and Tan Qiaoyu and Zhou Kaixiong and Zou Na and Hu Xia},
title = {Gradient rewiring for editable graph neural network training},
year = {2024},
booktitle = {NeurIPS}
}

@article{squash,
  title={Rewiring with positional encodings for graph neural networks},
  author={Gabrielsson Rickard Br{\"u}el and Yurochkin Mikhail and Solomon Justin},
  journal={Transactions on Machine Learning Research},
  year={2023}
}

@INPROCEEDINGS{new_related,
  author={Bourgerie Rémi and Zanouda Tahar},
  booktitle={2023 IEEE International Conference on Data Mining Workshops (ICDMW)}, 
  title={Fault Detection in Telecom Networks Using Bi-Level Federated Graph Neural Networks}, 
  year={2023},
}

\end{document}